\documentclass[journal, twocolumn]{IEEEtran}

\usepackage{subcaption}
\usepackage{xcolor}
\usepackage{caption}
\ifCLASSINFOpdf
  \usepackage[pdftex]{graphicx}
\else
\fi

\usepackage{amsmath}
\usepackage{amsfonts}

\usepackage{comment}
\usepackage{hyperref}
\usepackage{cite}

\begin{document}
%
\title{Rice Leaf Disease Detection: A Comparative Study Between CNN, Transformer and Non-neural Network Architectures}
%
%
%

\author{
    Samia Mehnaz\textsuperscript{[1]} \and Md. Touhidul Islam\textsuperscript{[2]} \\       \vspace{0.5cm} 

    \textsuperscript{[1]}Viqarunnisa Noon College, Dhaka, Bangladesh \\
    Email: samiamehnaz1003@gmail.com

    \textsuperscript{[2]}Department of Electrical and Electronic Engineering, \\
    Bangladesh University of Engineering and Technology, Dhaka, Bangladesh \\
    Email: 1806177@eee.buet.ac.bd
}

%


\maketitle

\begin{abstract}
In nations such as Bangladesh, agriculture plays a vital role in providing livelihoods for a significant portion of the population. Identifying and classifying plant diseases early is critical to prevent their spread and minimize their impact on crop yield and quality. Various computer vision techniques can be used for such detection and classification. While CNNs have been dominant on such image classification tasks, vision transformers has become equally good in recent time also. In this paper we study the various computer vision techniques for Bangladeshi rice leaf disease detection. We use the Dhan-Shomadhan -- a Bangladeshi rice leaf disease dataset, to experiment with various CNN and ViT models. We also compared the performance of such deep neural network architecture with traditional machine learning architecture like Support Vector Machine(SVM). We leveraged transfer learning for better generalization with lower amount of training data. Among the models tested, ResNet50 exhibited the best performance over other CNN and transformer-based models making it the optimal choice for this task. 
\end{abstract}

\begin{IEEEkeywords}
CNN, Transformer Architecture, SVM, Classfication, Rice Leaf Disease
\end{IEEEkeywords}

%
\IEEEpeerreviewmaketitle

\section{Introduction}
Rice is the staple food for the majority of the population in Bangladesh, contributing significantly to the country's economy and food security. With over 75\% of the agricultural land dedicated to rice cultivation, ensuring healthy rice crops is crucial for sustaining livelihoods and meeting national food demands. However, rice crops are highly vulnerable to various diseases such as bacterial blight, brown spots, and sheath blight, which can lead to significant yield losses if not addressed promptly. In Bangladesh, traditional methods of rice disease identification often rely on manual labor, where farmers visually inspect the fields for signs of disease. This process is not only subjective and error-prone but also energy-intensive, requiring considerable physical effort to cover large fields. Farmers must repeatedly monitor crops throughout the growing season, which demands significant time and labor resources. These challenges are exacerbated in rural areas where access to expert advice and modern tools is limited. The reliance on manual labor can delay timely interventions, leading to the rapid spread of diseases and substantial crop losses. 

The integration of modern technologies, such as image processing based detection offers a promising solution to these challenges. Automated disease detection systems can reduce the need for labor-intensive monitoring, providing farmers with precise, real-time diagnostics. This not only minimizes physical effort but also ensures faster and more accurate identification of diseases, allowing for timely and targeted treatments.

Over the years, various paradigms have emerged for image processing based classification. Early methods relied on techniques like Support Vector Machine (SVM), which provided a foundational approach to the task. With advancements in machine learning, Convolutional Neural Networks (CNNs) gained popularity for their superior ability to extract spatial features, leading to significant improvements in classification accuracy. More recently, transformer-based architectures have begun to dominate the field due to their ability to model long-range dependencies and achieve state-of-the-art results in many domains. In this research, we explored the relevance of all three paradigms—SVM, CNN, and transformer—in the context of Bangladeshi rice disease classification. We have experimented on the Dhan-Shomadhan dataset\cite{hossain2023dhanshomadhandatasetriceleaf} providing a comprehensive evaluation of the strengths and applicability of CNN, transformer, and non-neural network architecture like SVM in practical agricultural solutions.

\section{Literature Review}

The detection and classification of plant diseases using machine learning (ML) techniques has gained significant attention in recent years, particularly in agriculture where early disease detection is critical for crop management. Traditional machine learning models, such as Support Vector Machines (SVM), were initially employed to identify rice leaf diseases based on handcrafted features. In these early approaches, SVM was often used in conjunction with various feature extraction techniques such as color histograms and texture analysis \cite{sethy2020deep}. For instance, Sethy et al. \cite{sethy2020deep} developed a model using SVM for detecting different rice leaf diseases, including paddy leaf rot. While the model showed promising results, the reliance on handcrafted features limited its ability to capture complex patterns in the images.

With the increasing availability of large datasets, deep learning techniques have emerged as more powerful alternatives. Convolutional Neural Networks (CNNs), which are designed to automatically learn spatial hierarchies of features, have become the standard for image classification tasks. CNN-based models, such as ResNet and Inception, have demonstrated exceptional performance in various plant disease classification tasks \cite{tejaswini2022rice}. In \cite{tejaswini2022rice}, the authors applied CNN models such as VGG16, VGG19, ResNet, Xception for classifying plant diseases. This shift from traditional machine learning to deep learning marked a significant improvement in the ability to detect rice leaf diseases with minimal human intervention.

The next evolution in the application of neural networks to plant disease detection came with the advent of deeper and more sophisticated CNN architectures. For example, Krishnamoorthy et al. \cite{krishnamoorthy2021rice} employed a transfer learning approach using InceptionResNetV2 to diagnose rice leaf diseases, including bacterial blight, brown patches, and leaf blast. The authors leveraged a dataset of 5200 images and demonstrated the power of transfer learning for classifying rice leaf diseases with greater precision. Their work further emphasized the importance of pre-trained networks in dealing with small datasets, where training models from scratch may not be feasible.

In recent years, attention mechanisms have gained prominence as they allow models to focus on important features in the input data. One such architecture is MaxViT, a hybrid model combining the strengths of both Convolutional Neural Networks and Transformer-based attention mechanisms. MaxViT's ability to capture both local and global features through its multi-stage attention layers has shown promise in tasks requiring fine-grained feature extraction and contextual understanding \cite{tu2022maxvit}. This is particularly relevant in the context of rice leaf disease detection, where the intricate patterns of disease symptoms can benefit from global attention. MaxViT has shown impressive performance in various image classification tasks, making it a promising candidate for plant disease classification.

\section{Methodology}

This research presents a comparative analysis of rice leaf disease detection using CNNs, Transformer-based architectures, and non-neural network models. The methodology consists of data preprocessing, data augmentation, and detailed model implementation.

\subsection{Data Preprocessing}

The dataset comprised images of rice leaves exhibiting various disease symptoms. Each image was resized to a uniform dimension of $448 \times 448$ pixels to match the input requirements of the models. Pixel intensity normalization was applied to standardize the dataset, ensuring consistency in input image values.  

\subsection{Data Augmentation}

To increase data variability, enhance model generalization, and to avoid overfitting data augmentation techniques were utilized:
\begin{itemize}
    \item \textbf{Random Resize:} Images were randomly resized to simulate variations in scale.
    \item \textbf{Random Crop:} Random cropping introduced spatial variability in the input data.
    \item \textbf{Random Perspective:} Perspective transformations mimicked distortions typical in real-world imaging.
    \item \textbf{Random Gaussian Blur:} Blurring effects were applied to replicate noise conditions.
\end{itemize}

\begin{figure}
    \centering
    \includegraphics[width=0.8\linewidth]{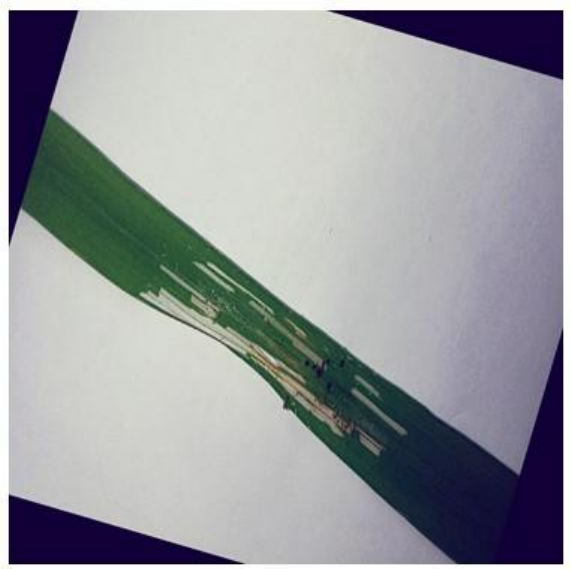}
    \caption{An Image with the all the augmentation transformation applied}
    \label{fig:image-w-aug}
\end{figure}

\subsection{Model Implementation}
Transfer learning was employed for all neural network models due to the relatively small size of the dataset and the challenges in acquiring large-scale labeled data in practical scenarios. All neural network models were pretrained on the ImageNet dataset, allowing them to leverage learned features and improve generalization. This approach significantly reduces the computational cost and training time while improving model performance. However, transfer learning was not applied to the SVM models, as they were either used directly or combined with features extracted from ResNet-based models.
\begin{figure}
    \centering
    \includegraphics[width=0.9\linewidth]{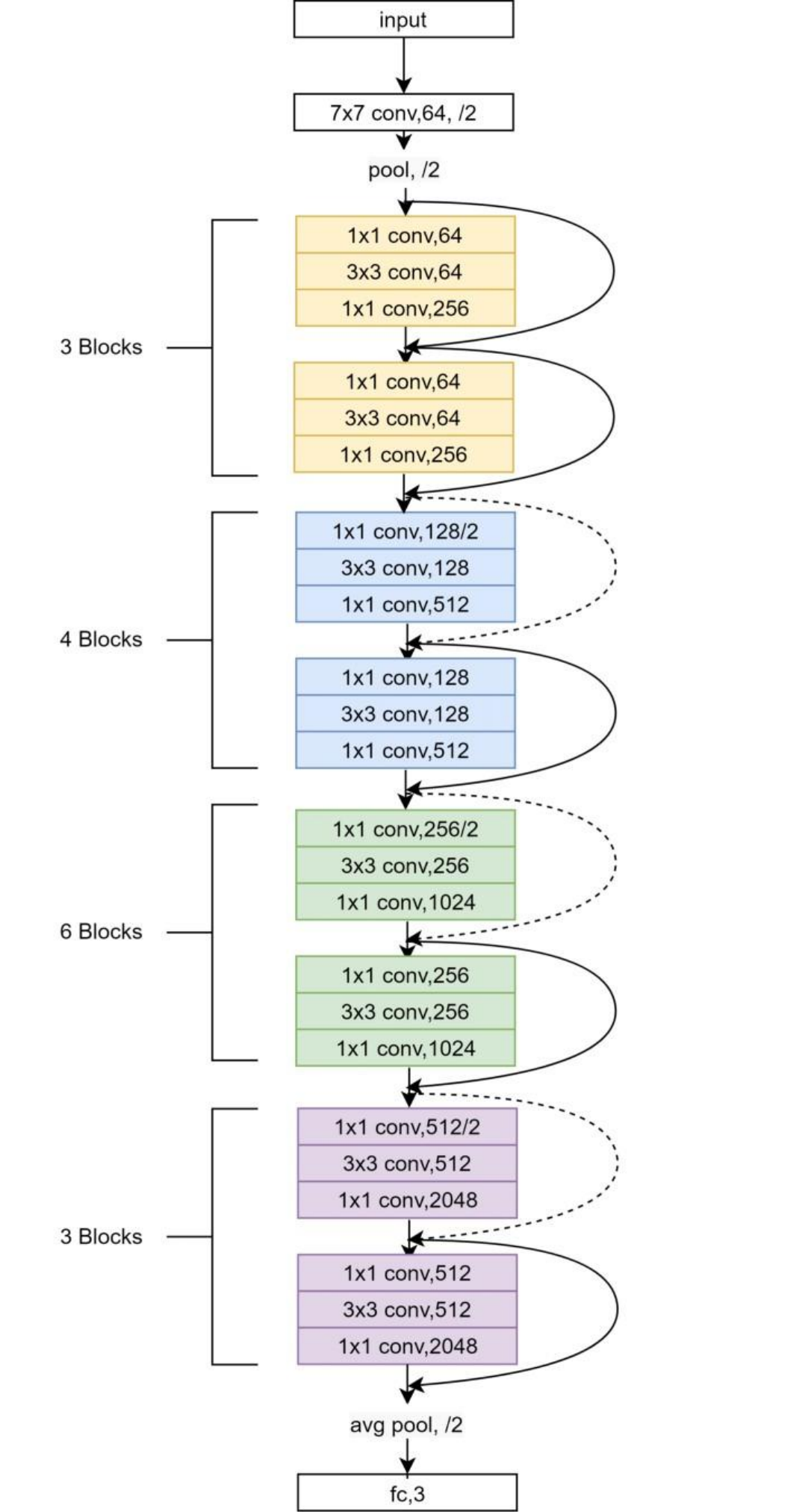}
    \caption{ResNet50 Model Architecture\cite{article}}
    \label{fig:resnet50-archi}
\end{figure}

\subsubsection{ResNet50}

ResNet50 is a deep convolutional neural network with 50 layers that employs residual learning to address the vanishing gradient problem commonly encountered in deep networks. By introducing skip connections, ResNet50 allows gradients to flow more effectively during backpropagation, making it possible to train very deep networks. Each residual block consists of identity mappings combined with convolutional operations, which enables the network to learn both low-level and high-level features efficiently. ResNet50 is widely known for its strong performance in image classification tasks due to its balance between depth and computational efficiency. In this study, ResNet50 was used to extract hierarchical features from rice leaf images, leveraging its ability to detect subtle patterns indicative of disease symptoms. Figure \ref{fig:resnet50-archi} shows the ResNet50 architecture. 

\subsubsection{ResNet152}

ResNet152 is a deeper variant of ResNet, consisting of 152 layers, and is designed to learn more complex and fine-grained features from input data. Similar to ResNet50, it incorporates residual connections, but the increased depth allows it to model intricate patterns in large datasets. The additional layers in ResNet152 make it particularly suitable for tasks that require capturing subtle variations in the data, such as differentiating between rice leaf diseases with similar visual symptoms. However, the increased depth comes with a higher computational cost. Despite this, ResNet152 has been shown to achieve superior accuracy in many classification tasks, making it a strong candidate for disease detection in this study.  

\subsubsection{Inception-V3}

Inception-V3 utilizes inception modules, which apply convolutional operations with multiple kernel sizes in parallel. This allows the network to capture features at various scales within a single layer. The architecture also incorporates auxiliary classifiers, which serve as intermediate supervision points during training, improving gradient flow and enhancing convergence. Additionally, Inception-V3 employs techniques such as factorized convolutions and label smoothing to improve training efficiency and model robustness. This architecture is particularly effective for handling complex datasets with diverse feature scales, making it suitable for rice leaf disease detection. In this study, Inception-V3 was applied to identify disease-specific patterns in the input images.

\begin{figure*}
    \centering
    \includegraphics[width=0.8\textwidth]{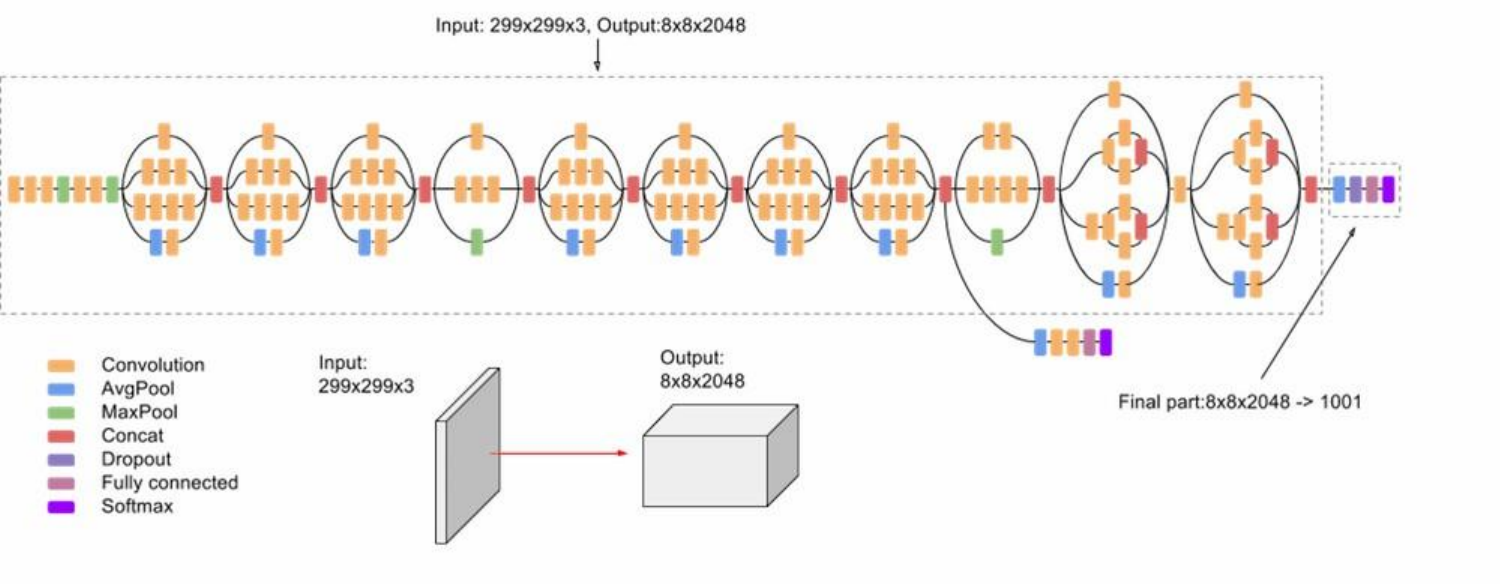}
    \caption{Inception-V3 Model Architecture\cite{7780677}}
    \label{fig:inceptionnet archi}
\end{figure*}

\subsubsection{MaxViT}

MaxViT is a Transformer-based architecture that combines local and global attention mechanisms for efficient feature extraction. It integrates convolutional layers for localized context and uses two attention mechanisms: Block Attention and Grid Attention.  

Block Attention divides the feature map into non-overlapping blocks and computes attention within each block, enabling the model to capture fine-grained details with reduced computational overhead. Grid Attention complements this by rearranging features into a grid-based structure, allowing for long-range dependency modeling across the image.  

A key component of MaxViT is the MBConv block (Mobile Inverted Bottleneck Convolution), which is used for efficient feature extraction. The MBConv block incorporates depthwise separable convolutions and squeeze-and-excitation (SE) layers to reduce computational complexity while enhancing feature representation. This block ensures that the model remains computationally efficient even for high-resolution images.  

\begin{figure*}[t]
    \centering
    \includegraphics[width=0.8\textwidth]{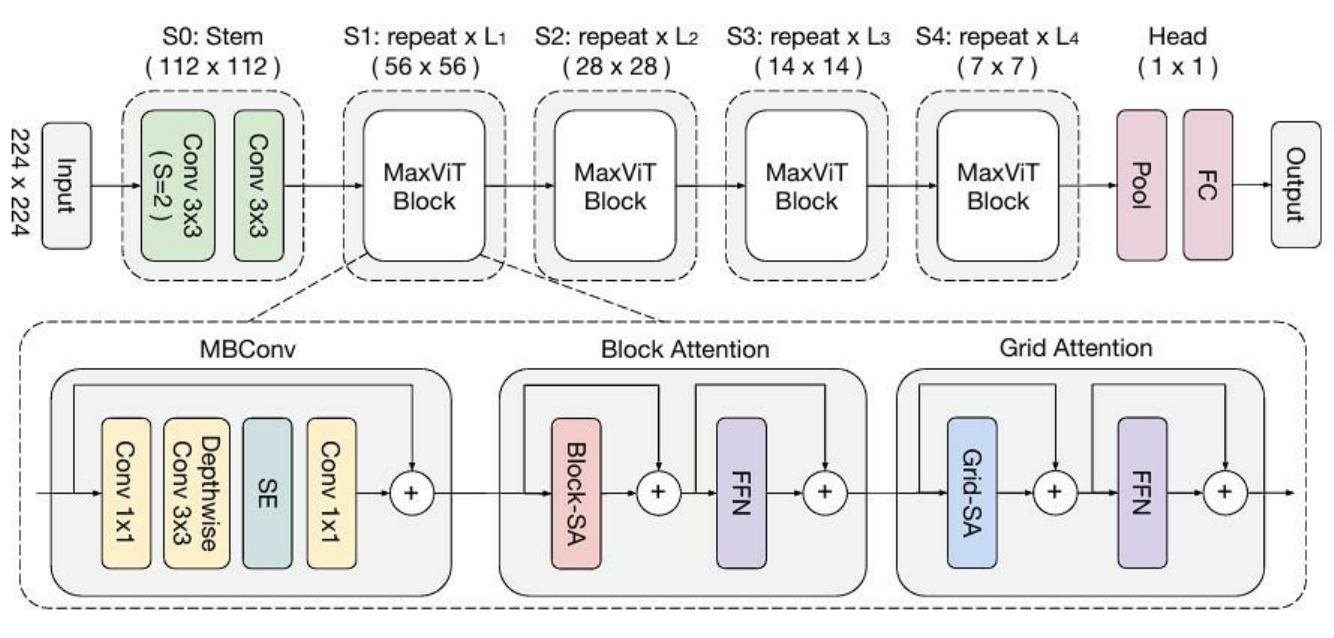}
    \caption{MaxViT Model Architecture \cite{tu2022maxvit}}
    \label{fig:maxvit-archi}
\end{figure*}

Each MaxViT block includes convolutional embedding, Layer Normalization, Multi-Head Self-Attention (MHSA) modules for Block and Grid Attention, and Feed-Forward Networks (FFNs) with skip connections for effective training. Figure \ref{fig:maxvit-archi}

MaxViT-L was employed in this study to evaluate the effectiveness of Transformer-based architectures in rice leaf disease detection, leveraging its ability to balance local detail extraction with global contextual understanding.

\subsubsection{Support Vector Machine (SVM)}

The SVM algorithm is a classical machine learning approach that separates data points of different classes using an optimal hyperplane. It is particularly effective for tasks where the data is linearly or near-linearly separable. In this study, SVM was applied as a baseline model for disease detection. The algorithm was trained on handcrafted features extracted from the dataset, providing a benchmark to compare the performance of deep learning models.

\subsubsection{SVM with ResNet Feature Extractor}

To leverage the strengths of both deep learning and classical machine learning, features were extracted from the penultimate layer of the ResNet50 model and used as input to an SVM classifier. This hybrid approach combines the robust feature extraction capabilities of ResNet50 with the classification power of SVM. By using deep, hierarchical features as input, the SVM classifier was able to achieve better performance compared to using raw image features. This method serves as a bridge between traditional and modern machine learning techniques, offering insights into their complementary strengths.

\section{Experimental Results}
\begin{table*}
    \centering
    \caption{Performance comparison of various models}
    \resizebox{0.7\textwidth}{!}{
    \begin{tabular}{p{3cm} c c c c}
        \hline
        \hline
        \textbf{Model} & \textbf{Validation Acc.} & \textbf{Recall} & \textbf{Precision} & \textbf{F1-Score} \\ 
        \hline
        \textbf{ResNet50\cite{he2015deepresiduallearningimage}} & \textbf{91.50\%} & \textbf{90.1\%} & \textbf{92.3\%} & \textbf{91.2\%} \\ 
        Inception-V3\cite{7780677} & 87.50\% & 86.23\% & 88.11\% & 87.50\% \\ 
        ResNet152\cite{he2015deepresiduallearningimage} & 90.00\% & 89.67\% & 91.2\% & 90.23\% \\ 
        MaxViT\cite{tu2022maxvit} & 75.00\% & 73.2\% & 76.4\% & 74.9\% \\ 
        SVM\cite{Cristianini2008} & 62.45\% & 60.21\% & 55.10\% & 55.43\% \\ 
        SVM+ResNet & 68.45\% & 61.1\% & 58.2\% & 59.89\% \\ 
        \hline
        \hline
    \end{tabular}
    }
    \caption*{\footnotesize{Note: All metrics have been macro averaged for the 5 classes.}}
    \label{tab:model_performance}
\end{table*}
We performed our experiments on the Dhan-Shomadhan dataset\cite{hossain2023dhanshomadhandatasetriceleaf}. The dataset contains images with white background and field background of 5 classes:  Rice Blast, Rice Tungro, Sheath Blight, Brown Spot, and Leaf Scald. All the images are collected from the region of Dhaka division of Bangladesh. Extensive experimentation was performed on a NVIDIA P100 GPU. Multistep learning rate was used and learning rate was decreased at optimal epochs. The dataset was split in an 80:10:10 ratio into train, test, and validation partitions. 

The evaluation metrics used listed below. Here TP, FP, TN, and FN respectively stands for True Positive, False Positive, True Negative, and False Negative. As we are dealing with multicalss classification, we worked with macro average of all the metrics.  
\begin{itemize}

    \item \textbf{Macro Average Precision:}
    \[
    \text{Precision}_{\text{macro}} = \frac{1}{N} \sum_{i=1}^{N} \text{Precision}_i
    \]
    Where:
    \[
    \text{Precision}_i = \frac{\text{TP}_i}{\text{TP}_i + \text{FP}_i}
    \]

    \item \textbf{Macro Average Recall:}
    \[
    \text{Recall}_{\text{macro}} = \frac{1}{N} \sum_{i=1}^{N} \text{Recall}_i
    \]
    Where:
    \[
    \text{Recall}_i = \frac{\text{TP}_i}{\text{TP}_i + \text{FN}_i}
    \]

    \item \textbf{Macro Average F1 Score:}
    \[
    F1_{\text{macro}} = \frac{1}{N} \sum_{i=1}^{N} F1_i
    \]
    Where:
    \[
    F1_i = 2 \cdot \frac{\text{Precision}_i \cdot \text{Recall}_i}{\text{Precision}_i + \text{Recall}_i}
    \]

    \item \textbf{Average Accuracy:}
    \[
    \text{Avg Accuracy} = \frac{1}{N} \sum_{i=1}^{N} \frac{\text{TP}_i}{\text{TP}_i + \text{FP}_i + \text{FN}_i + \text{TN}_i}
    \]

\end{itemize}

Extensive experimentation was conducted, testing various epochs, batch sizes, and learning rates to identify a robust model for accurate predictions. Table \ref{tab:model_performance} shows the experimental results for the various models. 

From the table, we can see that, ResNet50 achieved the highest macro f1-score of 91.2\%, followed by ResNet152 with 90.23\%, and Inception-V3 with 87.50\%. However, MaxViT demonstrated relatively lower performance, achieving a macro average f1-score of 74.9\%. What is notable is, all 3 best performing models are CNN-based, meaning that local context is more important for such leaf disease classification task. SVM underperformed severely with an f1-score of 55.43\%. Using ResNet for feature extraction of SVM and creating a hybrid model between SVM and ResNet improved the performance to an f1 score of 59.89\%. However that is far below the f1-score of the CNN networks. 

These results indicate that CNN models are superior for crop disease classification task. And traditional non-neural network machine learning based methods are ineffective on such task. ResNet50 is particularly effective for the rice crop disease detection, offering a balance between accuracy and training efficiency. 

\begin{figure}
    \centering
    \includegraphics[width=1\linewidth]{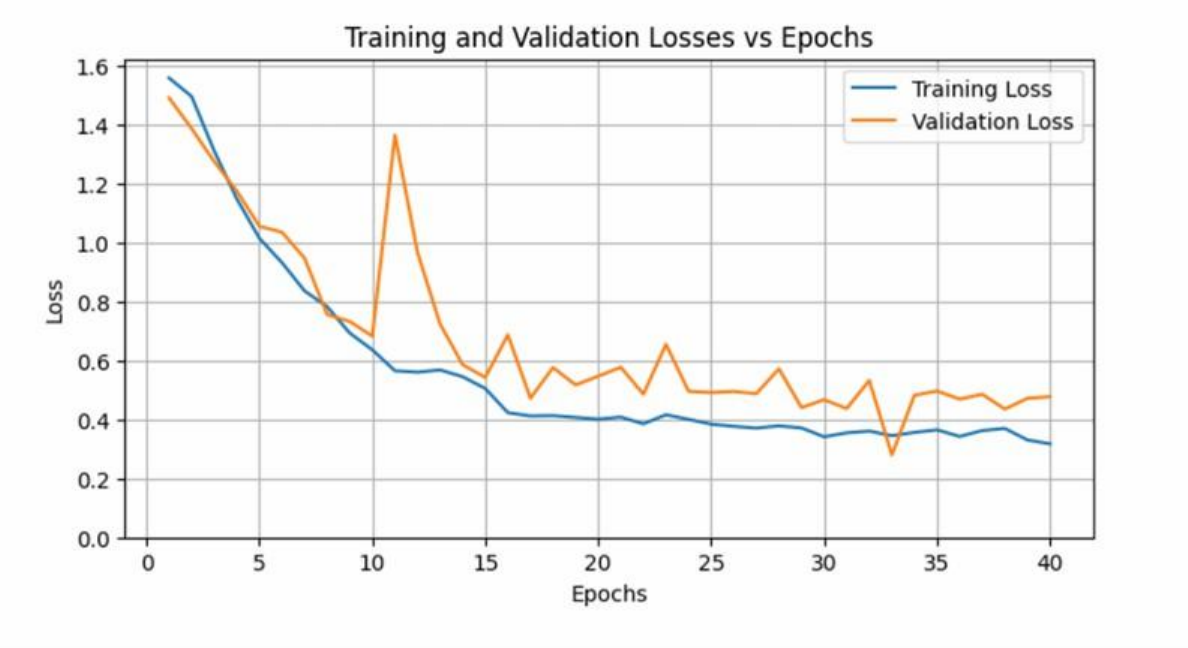}
    \caption{Epoch vs train and validation loss graph for ResNet50 training}
    \label{fig:resnet50-loss-graph}
\end{figure}

\begin{figure}
    \centering
    \includegraphics[width=1\linewidth]{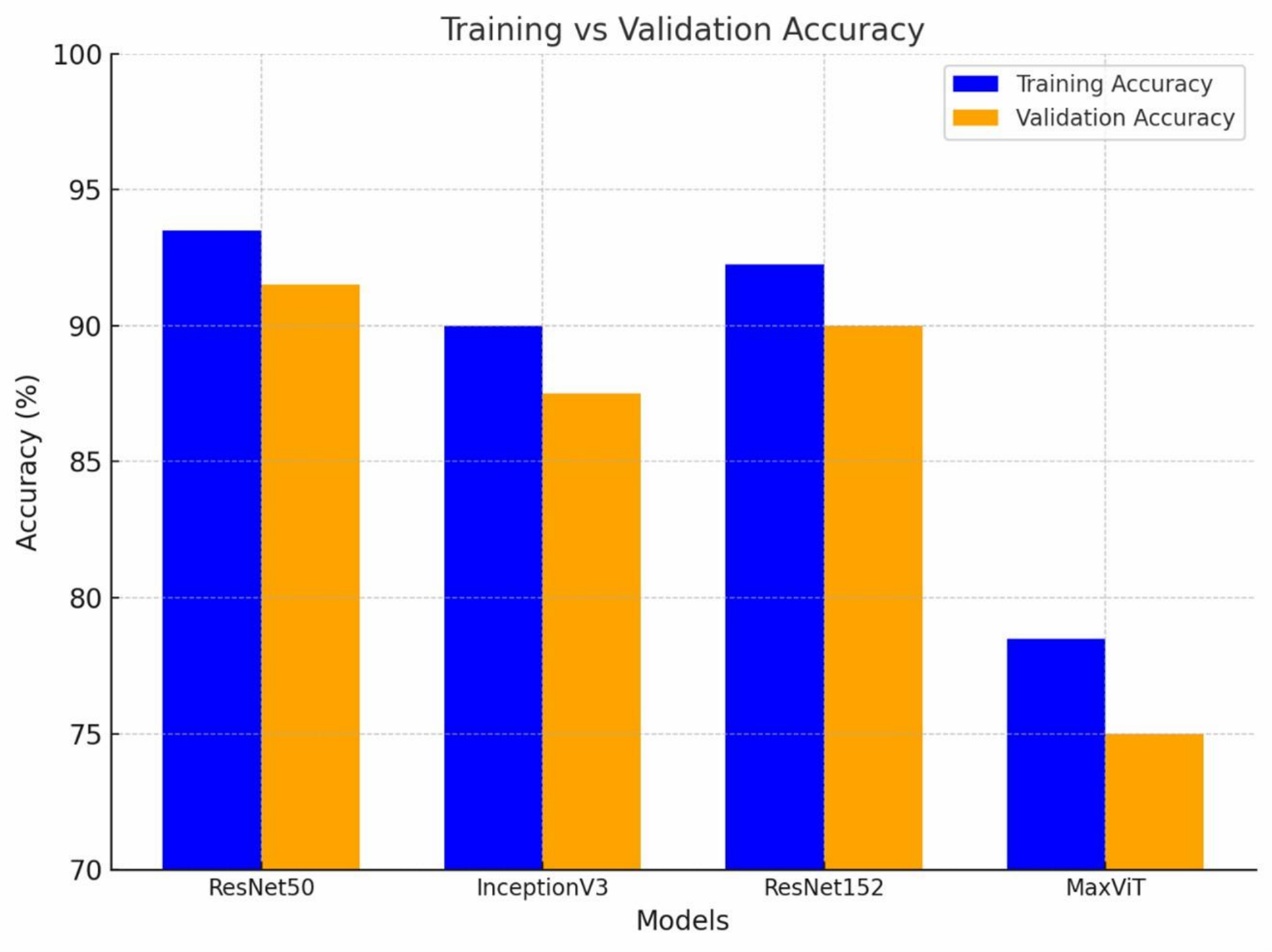}
    \caption{Comparison of macro averaged accuracy between the models}
    \label{fig:model-acc-comp}
\end{figure}

Figure \ref{fig:resnet50-loss-graph} shows the loss curve for the ResNet50 training. The validation loss converges to a stable value at around epoch 40 and an early stopping method has been adopted in training to overcome overfitting. Figure \ref{fig:model-acc-comp} shows comparison of accuracy between the neural network architectures.

\section{Conclusion}
Protecting crops in agriculture is a complex endeavor that demands a deep comprehension of the crops, pests, diseases, and weeds involved. In this reseach, we utilized deep learning models, particularly CNN architectures, to identify various rice diseases. The findings indicate that utilizing deeper models, such as ResNet152, typically enhances performance; however, ResNet50 provided a commendable balance between accuracy and computational efficiency. Our research highlights the capabilities of CNNs in automating the identification of rice diseases, simplifying the process for users to detect illnesses through visual input from cameras. By providing real-time feedback, farmers can take early action to safeguard their crops, thereby boosting productivity. Additional technologies like drones fitted with cameras and sensors can be integrated to improve the scalability and effectiveness of the disease detection system. This will further enable farmers to carry out preventive measures swiftly, ensuring healthier crops and enhanced yields.



%


\end{document}